# Graphical Abstract

**Fighting the disagreement in Explainable Machine Learning with consensus**

Antonio Jesús Banegas-Luna , Carlos Martínez-Cortés, Horacio Pérez-Sánchez

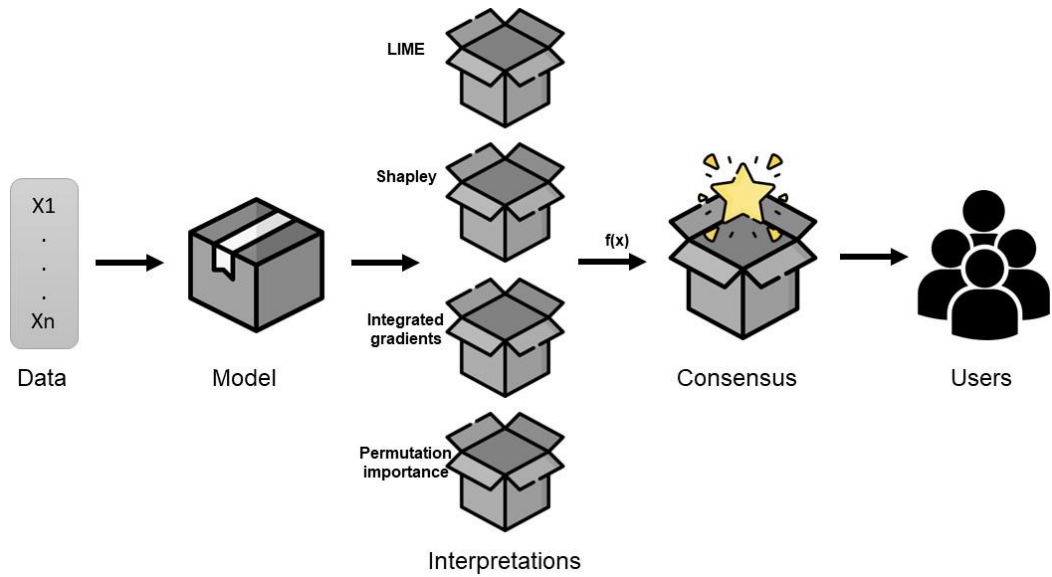

# Highlights

**Fighting the disagreement in Explainable Machine Learning with consensus**

Antonio Jesús Banegas-Luna, Carlos Martínez-Cortés, Horacio Pérez-Sánchez

- The interpretability of machine learning models is essential in critical contexts, such as medicine.
- There are several interpretability approaches that often disagree.
- The disagreement problem can be combated with consensus.
- A novel consensus function taking into account model accuracy, class probability and feature attribution has been developed.
- The proposed function leads to more consistent explanations than the others in four synthetic datasets.

# Fighting the disagreement in Explainable Machine Learning with consensus

Antonio Jesús Banegas-Luna [a], Carlos Martínez-Cortés[a], Horacio Pérez-Sánchez[a]

[a]*Structural Bioinformatics and High-Performance Computing (BIO-HPC), Universidad Católica de Murcia (UCAM), Campus de los Jerónimos, Guadalupe, 30107, Murcia, Spain*

**Abstract**

Machine learning (ML) models are often valued by the accuracy of their predictions. However, in some areas of science, the inner workings of models are as relevant as their accuracy. To understand how ML models work internally, the use of interpretability algorithms is the preferred option. Unfortunately, despite the diversity of algorithms available, they often disagree in explaining a model, leading to contradictory explanations. To cope with this issue, consensus functions can be applied once the models have been explained. Nevertheless, the problem is not completely solved because the final result will depend on the selected consensus function and other factors. In this paper, six consensus functions have been evaluated for the explanation of five ML models. The models were previously trained on four synthetic datasets whose internal rules were known in advance. The models were then explained with model-agnostic local and global interpretability algorithms. Finally, consensus was calculated with six different functions, including one developed by the authors. The results demonstrated that the proposed function is fairer than the others and provides more consistent and accurate explanations.

## 1. Introduction

Machine learning (ML) is a branch of artificial intelligence that aims to develop techniques to make computers learn. Thanks to ML, computers can perform certain tasks without being programmed for it. Its strong statistical basis makes ML a reliable approach to address complex tasks (e.g. pattern recognition, computer vision). Because of its accuracy in identifying patterns in data, ML models have been widely adopted in many fields of science, such as biology [1, 2], drug discovery [3, 4, 5], meteorology [6, 7, 8] and medicine [9, 10, 11].

ML models are frequently evaluated on the basis of their accuracy. Metrics such as AUC (Area Under the Curve), precision and recall are often used to measure the accuracy of classification models, while $R^2$, MSE (Mean Squared Error) and MAE (Mean Absolute Error) are common regression metrics [12]. However, some models still tend to be seen as black-boxes [13]. This opacity is frequently a problem in critical areas such as medicine or finance, where users need a deep understanding of the inner workings of models [14]. Such users are not only interested in accuracy but they also need to know how the model made the decision to gain knowledge. This need to interpret models has given rise to eXplainable Machine Learning (XML) [15]. The interpretation of models is often referred to in literature as explainability or interpretability, but some authors make a distinction between both terms [16]. This manuscript does not intend to discuss whether interpretability and explainability describe the same idea or how they should be defined [17]. Hence, aiming to simplify the explanations, both terms will be used indifferently.

There are several algorithms for explaining ML which can be classified according to many taxonomies [18]. For example, algorithms can be labelled as global or local depending on their ability to explain the entire dataset o each individual sample separately. From another perspective, algorithms can be classified according to their ability to focus on a specific model or to be model agnostic. Moreover, while some algorithms can be applied on different data types, others are specifically designed to deal with certain types of data.

Despite its several advantages and usefulness, XML has some limitations too [19]. The first problem is defining what an explanation is. In this manuscript, an explanation is defined as a collection of input features that carries the most weight in the model's predictions. Consequently, our goal will be to identify such features. Another problem is related to the accuracy of the explanations [20]. In this respect, it is important to bear in mind that explanations can be inaccurate, as well as models are. Finally, it should be highlighted that, since each interpretability algorithm is based on a different idea, they may disagree [21, 22]. The problem of disagreement often leads to contradictory and fuzzy explanations, which makes interpretability fail. With regard to disagreement, it is crucial to find an effective way to combine the explanations obtained from interpretability. In this sense, consensus can be a helpful alternative.

A variety of consensus functions could be applied to achieve a consistent overall explanation of the models. The arithmetic mean is probably the most straightforward approach, but other means (e.g. harmonic, geometric) could be considered too. In addition, other approaches based on the number of appearances of the features among the most attributed ones could also be a good choice. Unfortunately, the less accurate the model, the less reliable its predictions and, in consequence, its explanations. Thus, other factors, such as model accuracy, should be included in the calculation of consensus.

This manuscript assess a diversity of consensus functions to explain some ML models. As the evaluated functions mainly focus on the attribution given to the features and omit the additional factors mentioned above, a novel consensus function is proposed. Section 2 describes the models, datasets, interpretability algorithms and consensus functions that have been used in the experiments. Next, section 3 summarizes the metrics obtained after training the models and the explanations given by each function for these predictions. In section 4 the performance of the consensus functions is extensively evaluated and justified. Finally, the main conclusions and future works are presented.

## 2. Materials and Methods

This section describes the datasets that were created for experiment, the models with which they were trained, the interpretability algorithms used for explaining the internal workings of the models and the functions that made consensus of the explanations. In addition, a novel consensus function proposed by the authors is described.

*2.1. Datasets*

Aiming to analyze the disagreement in classification and regression problems, four synthetic datasets were created. The main reason to opt for syn-thetic datasets rather than for toy datasets is that the input features involved in the calculation of the output class has to be known. Otherwise, the accuracy of the explanations could not be validated.

Table 1: Details of the datasets used in this work.

| Name | Type | Samples | Features | Explanation |
|---|---|---|---|---|
| DS1 | Classification | 2000 | 20 | F2, F3, F9, F17 |
| | *if(F2 ∗ F3)/F9 < F17 then 0 else 1* | | | |
| DS2 | Classification | 1500 | 75 | F5, F25, F55 |
| | *if (F55³ + F5² − F25 < 0) then 0 else 1* | | | |
| DS3 | Regression | 2500 | 60 | F1, F56, F58, F60 |
| | *sin(F60) + cos(F58) + tanh(F56) + F1* | | | |
| DS4 | Regression | 2000 | 30 | F19, F21, F24, F26 |
| | *F19⁴ − F21³ + F24² − F26* | | | |

Since SIBILA tool [23] was used to automate and accelerate the calculations, the classification datasets were adapted to resolve binary classification problems. Concerning regression, the output was calculated on the basis of some of the input features by applying different linear and non-linear functions. Additionally, a number of noise features were added to each dataset. This was done to verify if the interpretability algorithms focused on the most relevant features or they were disturbed by the noise instead.

The datasets contained between 1500 and 2500 samples and 20 and 75 features. These sizes were chosen to make the effect of noise really effective but avoid too time consuming calculations. All the features were created with random numbers between 0 and 1 following a uniform distribution. That way, we simulate a normalized dataset improving model accuracy and, consequently, leading to better explanations.

Table 1 summarizes the sizes of each dataset along with the rules implemented. The *Explanation* column lists the features that interpretability algorithms were expected to identify. Under each dataset, the formula implemented to calculate the output value is described.

*2.2. Machine Learning models*

The analyzed algorithms were used to interpret the predictions of different ML models. For this purpose, a collection of models was selected with the only restriction that they should not be interpretable by themselves. Thus, models such as decision trees, which explain their models graphically, and other rule-based models (e.g. RuleFit [24], RIPPERk [25]) were not included in the study.

On the contrary, the selected models were k-nearest neighbours (KNN), random forest (RF), support vector machines (SVM), gradient boosting machines (XGB) and artificial neural networks (ANN). They are a representative collection that includes non-linear, ensemble and black-box models.

*2.3. Interpretability algorithms*

A representative set of interpretability algorithms were chosen for this experiment. Both global and local model-agnostic approaches were taken into consideration. It is worth noticing that global methods represent the contribution of a feature to model decisions with a single numerical value. That value is typically called *attribution*. Often, in global methods, the attribution of a feature is the average of the individual attributions of the feature obtained in each of the individual samples. However, local approaches return the individual attribution of each feature in each of the samples.

In this work, permutation importance [26] and random forest based feature importance [27] represented the global algorithms, while LIME [28], Shapley values [29, 30], integrated gradients [31] and counterfactuals [32] the local ones. All the algorithms used in this work are model-agnostic, which means that they can be used to explain any type of model.

*2.4. Consensus functions*

Consensus can be calculated in several ways. Probably, the most straightforward is the arithmetic mean ($A_{mean}$) of the attributions given to each feature. This is a very simple and effective way to combine the explanations of the different interpretability approaches into a single result. However, the effect of noise or low accuracy models might impact negatively its results.

Although the arithmetic mean is probably the most frequently used function when combining data from different sources, there are other types of means such as the harmonic ($H_{mean}$) or the geometric ($G_{mean}$) ones. The harmonic mean is calculated as the opposite of the arithmetic mean applied to the inverse of the elements (Eq. 1). In some cases, it may be more representative than the arithmetic mean, but it does not work with null or negative values. On the other hand, the geometric mean is $n^{th}$ root of the product of all the attributions (Eq. 2). It is less sensitive to extreme values than the arithmetic mean but its calculation is more complex. Furthermore, as well as the harmonic mean, it does not work correctly with non-positive values.

$$H_{mean} = \sum_{i=1}^{n} \frac{n}{\frac{1}{x_i}} \qquad (1)$$

$$G_{mean} = \sqrt[n]{\prod_{i=1}^{n} x_i} \qquad (2)$$

A limitation due to the usage of attribution scores is that not all interpretability algorithms use the same range of values, which can make it difficult to compare the explanations of different algorithms. To overcome this issue, the attributions can be normalized. The alternative to the normalization is to sort the features by attribution in descending order and take the relative position of each feature as the score to make consensus with. Therefore, the lower the position, the more important the feature.

Another option based on the relative position of the features is the voting approach, which is based on random forest. In its shortened version, this function sorts the features by attribution in descending order, takes the $N$ most attributed features and counts how many times each feature is between the first and the $N^{th}$ position. Finally, the features that appear most frequently in the top N are assumed to be the most important. As well as the previous function, voting omits the attributions assigned to the features.

In recent years, many studies about the application of ML/DL models to the field of data fusion have been published [33, 34, 35]. However, due to the complexity of the proposed models and the expensive computational requirements they could bring, it was decided to try simple functions first.

## 2.5. Development of a novel consensus function

The functions described in the previous section omit some parameters that we consider crucial to perform consensus correctly. Because of this, we decided to develop a novel function including these factors.

First, the interpretability approaches used in this work do not set their attributions in the same range (e.g. between 0 and 1) what makes difficult the comparison between them. In consequence, our novel consensus function will normalize all the attributions in the range [0, 1], using the min-max approach, to perform a fair evaluation (Eq. 3).

$$attr' = \frac{attr - min(attr)}{max(attr) - min(attr)} \tag{3}$$

Secondly, two important parameters are often omitted: the accuracy of the model and the class probability of the samples. The model accuracy should be taken into account when performing consensus because the reliability of the explanations can depend on it [36, 37]. An inaccurate model will fail most of its predictions and, consequently, the explanations obtained from it will be inaccurate or wrong too. Therefore, if we omit this parameter we would be considering that all the explanations are always correct, which is not true. Model accuracy can be modelled as a function $f(AUC)$ which has its minimum when $AUC = 0.5$ and its maximum when $AUC = 1$ or $AUC = 0$. If $AUC = 0.5$ means the model behaves like a random classifier, thus it is the least reliable model that can be obtained. On the contrary, when $AUC = 1$ the model behaves like a perfect classifier and when $AUC = 0$ it is always mistaken, thus all its predictions just need to be inverted.

The impact of model accuracy in the attribution of a feature, $\alpha$, is graphically shown in Fig. 1 and calculated in Eq. 4.

$$\alpha = 4(AUC^2 - AUC) + 1 \tag{4}$$

Taking into account model accuracy may be enough for global methods, but not for the local ones, which explain each sample individually. In this case, not all the samples are predicted with the same certainty. For example, a sample which is predicted with a probability of 0.5 is the result of a random decision. However, another sample which was classified with a probability of 0.99 is a very reliable prediction. Hence, this is an essential parameter when assessing local methods. The impact of class probability is represented by β in Eq. 5 and it is modeled exactly as α. While a class probability of *proba* = 0.5 represents a random decision for the sample, a probability of *proba* = 1 means the sample was undoubtedly classified in its class, and when *proba* = 0 the sample is classified as the opposite class. Thus, β is calculated as indicated in Eq. 5.

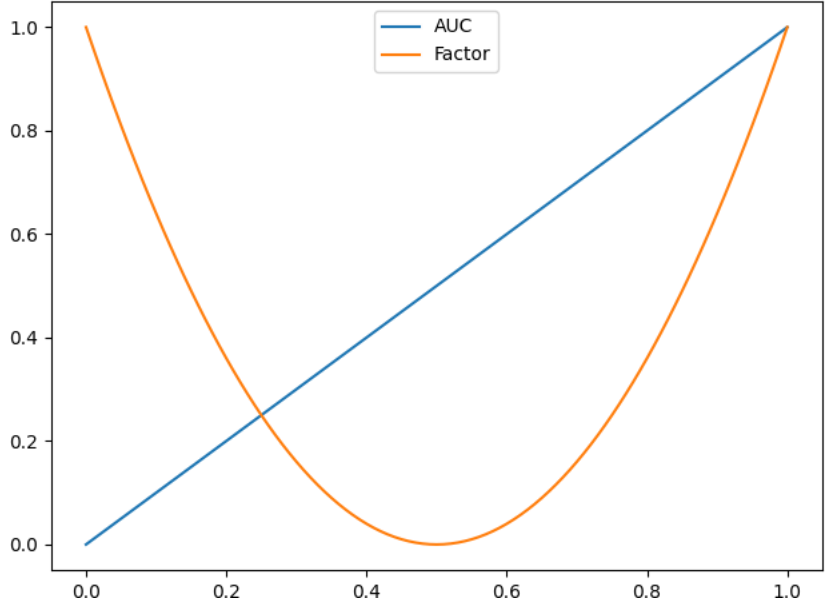

Figure 1: Behavior of AUC and its impact on consensus.

$$β = 4(proba^2 - proba) + 1 \qquad (5)$$

Joining α and β in one single equation, we will have a first approach to our novel function. As shown in Eq. 6, the global combined attribution can be depicted as a function of the normalized initial attribution multiplied by the impact of the model accuracy and class probability on the explanation.

$$f(attr) = attr' * α * β \qquad (6)$$

Where $attr'$ is the normalized attribution given to the feature, α is the effect of the model accuracy and *β* is the effect of the class probability.

There is a final remark to consider when evaluating consensus. Global interpretability methods return a single attribution value per feature, which is inferred from the attributions of all the individual samples. On the contrary, local methods return a single value per sample. Therefore, if both values were combined in one single equation, local methods would have more importance simply because they contribute with more values. To overcome this problem, the attributions of the local methods have to be divided by the number of samples, so that each local attribution contributes in the same proportion as the global ones.

Having all these considerations in mind, the consensus can be performed as the combination of two formulae: one for global methods (Eq. 7) and another one for local methods (Eq. 8).

$$f_{global}(attr) = attr' * \alpha \tag{7}$$

$$f_{local}(attr) = \frac{attr'}{N} * \alpha * \beta \tag{8}$$

## 3. Results

This section details the results obtained during the training of the models and the different consensus functions after interpretability.

### 3.1. Model training

Aiming at obtaining reliable results, each model was trained 50 times with each dataset. Next, the most accurate instance of each model was selected for interpretation. *AUC* in classification problems and $R^2$ in regression were the metrics chosen for evaluation. Table 2 lists the metrics obtained from each model with the four datasets.

### 3.2. Consensus

To assess the accuracy of the consensus functions, they were tested on the four synthetic datasets. The output of the datasets was calculated manually with a different rule in each case. Therefore, the features ruling the output were known beforehand. Those features are summarized in Table 1 and they were expected to be identified by the consensus functions. Based on the fact that all the datasets can be explained with at most 4 features, only the five most attributed features were analyzed for assessment purposes. Then, our goal was to find all the relevant variables in this subset.

Table 2: Performance metrics obtained after training 50 times each model.

| | Classification[a] | | Regression[b] | |
|---|---|---|---|---|
| **Model** | **DS1** | **DS2** | **DS3** | **DS4** |
| ANN | 0.914±0.01 | 0.775±0.02 | 0.981±0.00 | 0.892±0.01 |
| KNN | 0.784±0.02 | 0.595±0.03 | 0.395±0.02 | 0.517±0.03 |
| RF | 0.947±0.01 | 0.818±0.02 | 0.951±0.00 | 0.951±0.01 |
| SVM | 0.833±0.03 | 0.756±0.07 | 0.986±0.00 | 0.900±0.01 |
| XGB | 0.931±0.01 | 0.745±0.02 | 0.965±0.00 | 0.966±0.00 |

[a] Measured in AUC, [b] Measured in $R^2$

Figures 2 to 7 show the hits of each consensus function when detecting the expected features of the datasets. Each bar represents the feature in the $N^{th}$ position where they are sorted by attribution in descending order. Green bars represent that the feature at that position is among the expected ones, while red bars means a fail in the explanation due to noise. In consequence, it was desired to have all the green bars in the first $N$ positions. Additionally, the length of the bar represents the attribution score given to each feature. In the case of voting, it depicts the number of times that variable appeared among the most important. And, concerning the ranking function, each bar shows the average position of the feature when sorted by attribution. Therefore, large green bars and short red bars were expected, except in the case of the ranking function where short green bars and large red bars were expected. As can be observed in the figures, the proposed function is the one obtaining the best results.

## 4. Discussion

According to the results, the harmonic and geometric mean are clearly the worst approaches. They performed similarly and rarely identified the expected variables. The harmonic mean always failed in its explanations. This type of mean is usually the preferred choice to average times, distances and velocities [38], that is continuous values changing over time. This is not the case of attributions, whose values are continuous but deterministic. Since it is calculated as the opposite of the arithmetic mean, it seems logical to think that if the arithmetic mean performs well, the harmonic mean does not.

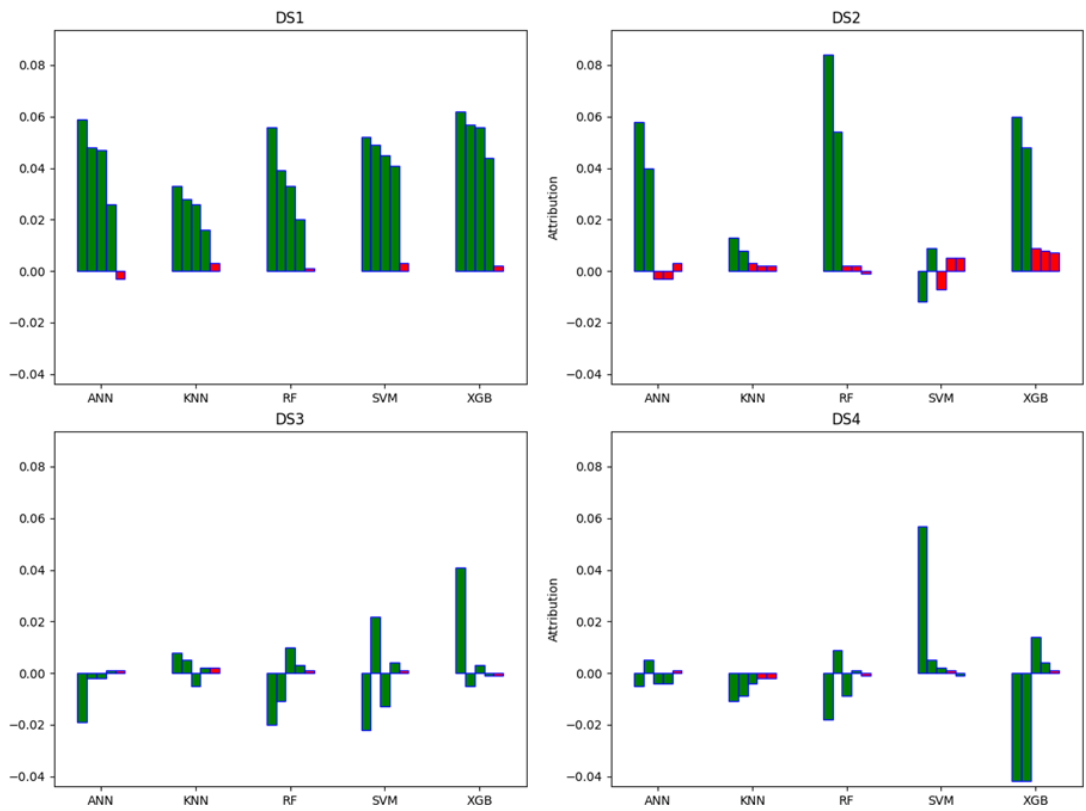

Figure 2: Matches of the expected features with the arithmetic mean.

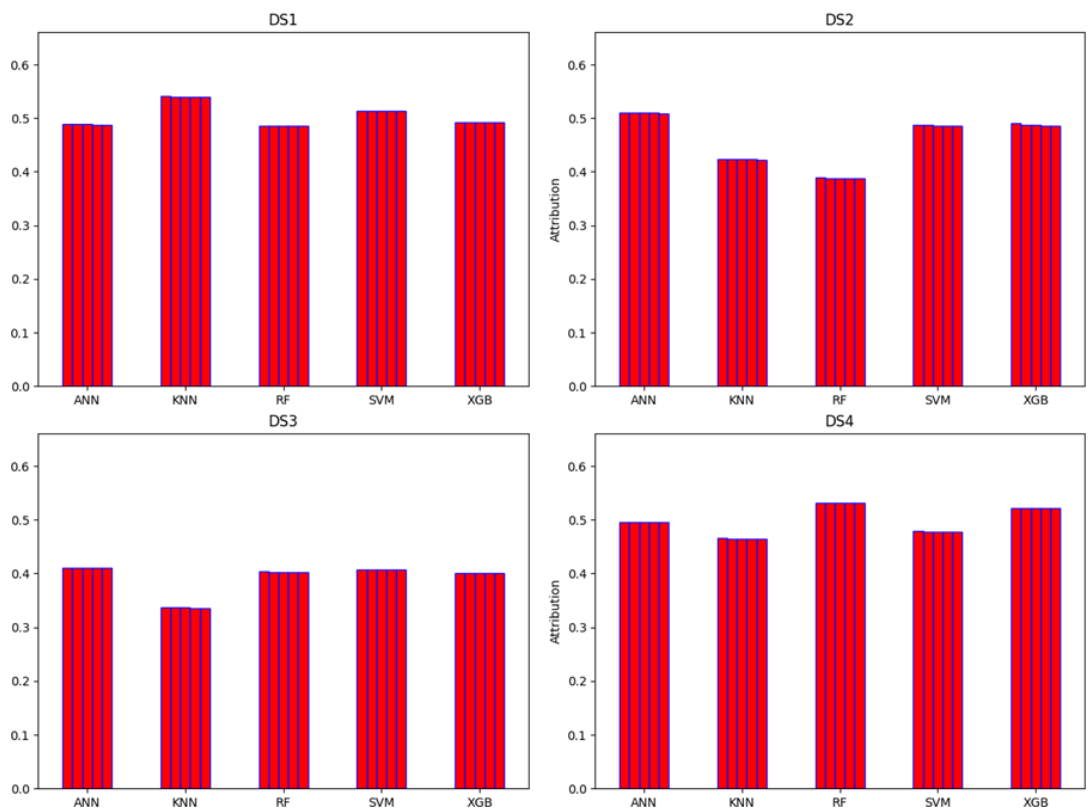

Figure 3: Matches of the expected features with the harmonic mean.

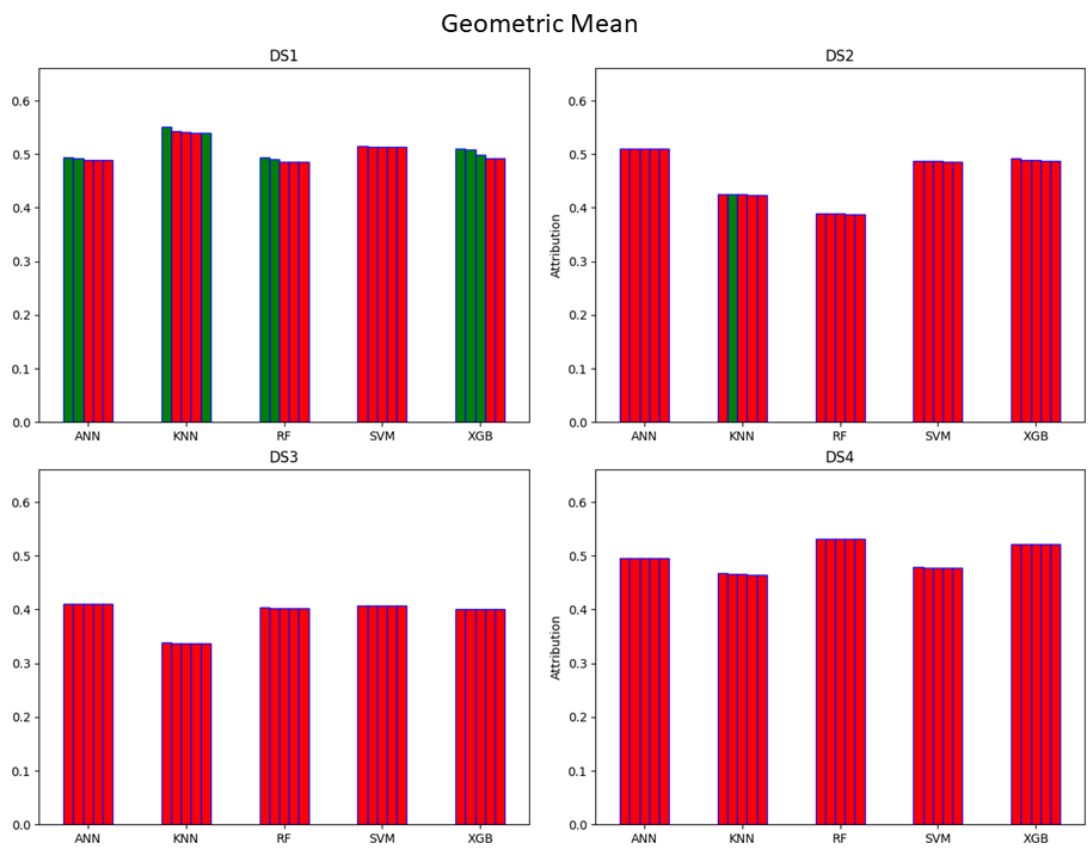

Figure 4: Matches of the expected features the geometric mean.

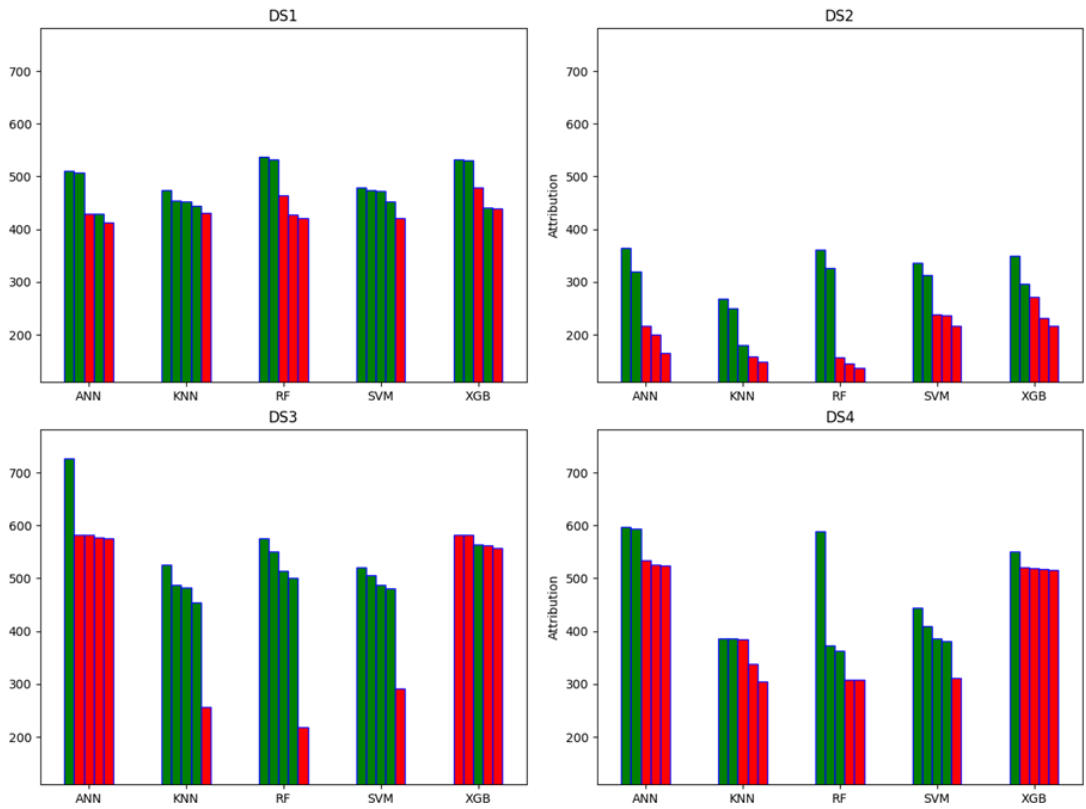

Figure 5: Matches of the expected features with the voting function.

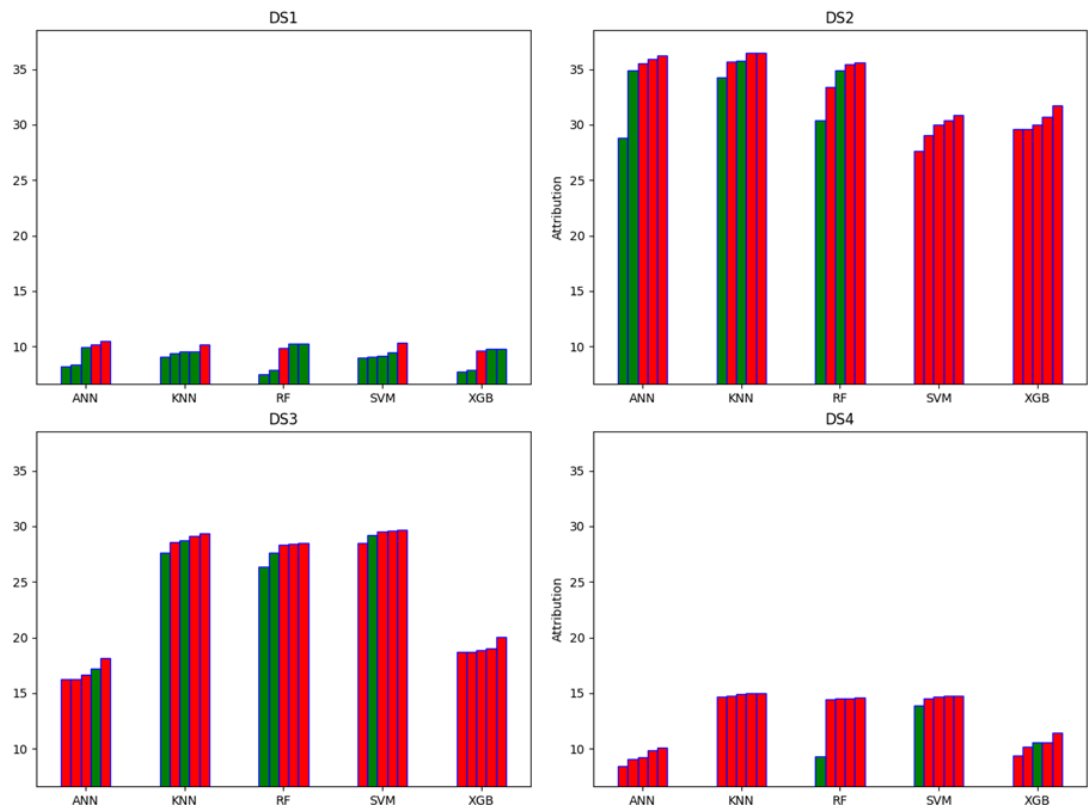

Figure 6: Matches of the expected features with the ranking function.

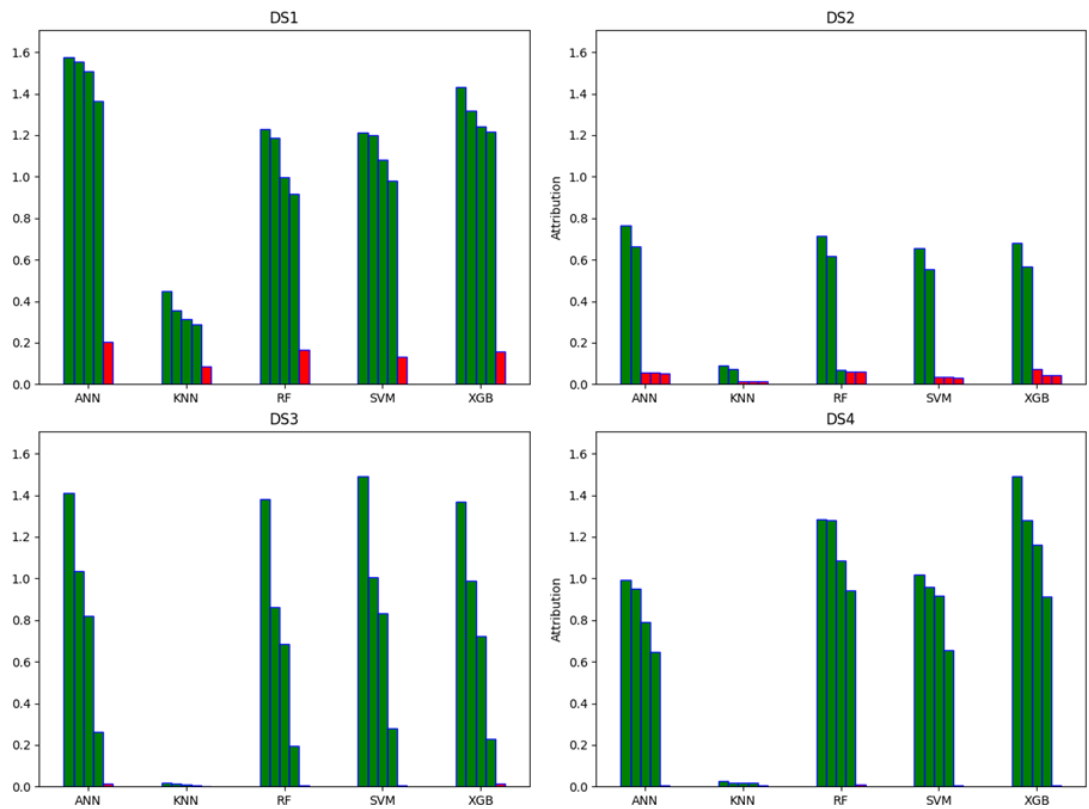

Figure 7: Matches of the expected features with the proposed function.

The geometric mean was only able to explain a few cases in DS1 and one feature in DS2 for the KNN model. But it failed all the explanations in DS3 and DS4, which seems to indicate that this is not a good function for regression problems. As shown in Table 2, the selected models achieved higher accuracy in DS1 than in DS2. Therefore, the geometric mean could be a feasible alternative for the explanation of highly accurate classification models. Nevertheless, its performance as a consensus function was really far from the expected.

The average ranking of the features is very easy to calculate but unfortunately has many limitations. First, this function ignores the attribution values, thus it is not able to tell whether a feature is much more relevant than another. Moreover, some interpretability algorithms, such as permutation importance, tend to focus on a few features and do not attribute anything to the others. As a result, many features have the same attribution and the order between them cannot be unambiguously determined. The results demonstrate its poor capacity to explain ML models. Only in 2 out of 20 tests it was able to identify the most relevant features in the dataset. In addition, the relative position of the relevant features was frequently high, which means that they are often considered to be a low relevance. Although this issue stems from the interpretability algorithms, a robust consensus function is expected to cope with this situation. Furthermore, there are no major differences between the position of relevant and noise features. This makes it difficult to determine which are really important when an unexplored dataset has to be analyzed.

Similar issues were observed in the voting function. As well as the average ranking, this function is very sensitive to noise because when a feature with low relevance is consistently placed next to the important ones, it will obtain several votes. Therefore, such feature will indirectly gain a lot of importance. Furthermore, voting, as well as the ranking function, do not consider the sign of the attributions. This is a crucial factor to determine if a feature contributes positively or negatively in a decision. Unfortunately, these two approaches ignore that information, consequently, all their attribu-tions are positive. Despite its limitations, the voting approach outperformed other functions (harmonic mean, geometric mean and ranking) and, in some cases, was able to give the relevant features significantly higher values than noisy ones. However, its performance was not good enough to be considered a reliable option.

The results show that the arithmetic mean is clearly the most accurate function among the "classical" ones. This function performed nearly perfect in all the datasets. Such a good behaviour may be explained by the fact that it is able to handle null and negative values, and the more samples explained the more it minimizes errors. However, two limitations were observed. The first problem is that the arithmetic mean ignores the accuracy of the model. This means that the explanations coming from a random model will receive the same attention as the ones coming from a perfect model. This lead to inaccurate results because the quality of the explanations is directly connected with the accuracy of the model. In addition, not all the interpretability algorithms handle attributions in the same range. As a result, the algorithms setting higher attributions will have more impact than the others. Despite all these problems, the arithmetic mean showed very good performance.

To overcome all the aforementioned issues, a novel consensus function was developed. Our function, which takes into consideration the model accuracy and class probability, outperformed the arithmetic mean. Additionally, the attributions were normalized between 0 and 1 to cope with values in different scales. In the first dataset, both functions identified the expected features in all the models. However, in the novel function, the difference between the attributions of the relevant and noisy features was higher than in the arithmetic mean. This is a key point in order to help the users to decide unequivocally which are the most important input features. In the second dataset, the novel function outperformed the arithmetic mean again. It was able to identify the three expected features in the RF model, while the arithmetic mean did not identify $F$ 55. The results of the third dataset were nearly the same except that the novel function highlighted the relevant features more than the mean. Finally, in the fourth dataset, again, there was a slight difference between both functions. It is worth mentioning that the KNN model obtained a very low accuracy ($AUC = 0.517$), in consequence, it explanations were inaccurate what made difficult to identify the features involved in its formula. As expected, the arithmetic mean could not detect the four features. However, our novel function was able because it penalized the explanations of this bad model. This finding confirms that weighting the model accuracy and the class probability helps consensus to obtain good global explanations of the models.

## 5. Conclusions

Interpretability is a crucial point to understand the inner workings of black-box ML models. Many approaches have aroused in the last years to explain the way the models take decisions. Although the idea behind all of them is to calculate the contribution of each input feature in a numerical way, they adopt different algorithms to do it. In order to overcome the disagreement among the algorithms, consensus may be a good alternative. Consensus can be implemented in several ways, such as the mean, the relative position of the features or the number of times they appear between the most relevant ones.

In this work, five consensus functions have been evaluated on four synthetic datasets. The results proved that all had limitations, therefore a novel consensus function was proposed. That novel function took into consideration the model accuracy, the class probability of the samples and the normalization of the attributions as essential factors to calculate an overall explanation of a model. The proposed function outperformed the others because it was able to detect the features used to calculate the output in the majority of the cases. The findings observed in this work demonstrate that consensus need to take into account many factors to be really accurate. Having an effective consensus function could help scientists to explain the predictions of their models to the final users, which is essential in fields such as medicine.

Future works will focus on the application of consensus to real-life datasets whose internal rules are completely unknown. In addition, our novel function should be tested with multiclass datasets and in the interpretation of more complex models, such as convolutional neural networks or time series. Finally, the set of consensus functions could be extended. In this regard, the development of an ML-based consensus function could be an alternative.

**Funding**

This work has been funded by grants from the European Project Horizon 2020 SC1-BHC-02-2019 [REVERT, ID:848098]; Fundación Séneca del Centro de Coordinación de la Investigación de la Región de Murcia [Project 20988/PI/18]; and the Spanish Ministry of Economy and Competitiveness [CTQ2017-87974-R].

**Declaration of Competing Interest**

The authors declare no competing interest.

**Acknowledgements**

This work has been funded by grants from the European Project Horizon 2020 SC1-BHC-02-2019 [REVERT, ID:848098]; Fundación Séneca del Centro de Coordinación de la Investigación de la Región de Murcia [Project 20988/PI/18]]; and the Spanish Ministry of Economy and Competitiveness [CTQ2017-87974-R]. Supercomputing resources in this work were supported by the Poznan Supercomputing Center's infrastructures, the e-infrastructure program of the Research Council of Norway, and the supercomputing centre of UiT—the Arctic University of Norway, the Plataforma Andaluza de Bioinformática at the University of Málaga, the supercomputing infrastructure of the NLHPC (ECM-02, Powered@NLHPC), and the Extremadura Research Centre for Advanced Technologies (CETA-CIEMAT), funded by the European Regional Development Fund (ERDF). CETA-CIEMAT is part of CIEMAT and the Government of Spain.

[15] L. H. Gilpin, D. Bau, B. Z. Yuan, A. Bajwa, M. Specter, L. Kagal, Explaining explanations: An overview of interpretability of machine learning, in: 2018 IEEE 5th International Conference on data science and advanced analytics (DSAA), IEEE, 2018, pp. 80–89.

[16] A. Nandi, A. K. Pal, Interpreting machine learning models: Learn model interpretability and explainability methods, Springer, 2022.

[17] M. Krishnan, Against interpretability: a critical examination of the interpretability problem in machine learning, Philosophy & Technology 33 (3) (2020) 487–502.

[18] P. Linardatos, V. Papastefanopoulos, S. Kotsiantis, Explainable ai: A review of machine learning interpretability methods, Entropy 23 (1) (2020) 18.

[19] C. Rudin, C. Chen, Z. Chen, H. Huang, L. Semenova, C. Zhong, Interpretable machine learning: Fundamental principles and 10 grand challenges, Statistic Surveys 16 (2022) 1–85.

[20] J. Zhou, A. H. Gandomi, F. Chen, A. Holzinger, Evaluating the quality of machine learning explanations: A survey on methods and metrics, Electronics 10 (5) (2021) 593.

[21] S. Krishna, T. Han, A. Gu, J. Pombra, S. Jabbari, S. Wu, H. Lakkaraju, The disagreement problem in explainable machine learning: A practitioner's perspective, arXiv preprint arXiv:2202.01602 (2022).

[22] M. Flora, C. Potvin, A. McGovern, S. Handler, Comparing explanation methods for traditional machine learning models part 1: An overview of current methods and quantifying their disagreement, arXiv preprint arXiv:2211.08943 (2022).

[23] A. Banegas-Luna, H. Pérez-Sánchez, Sibila: High-performance computing and interpretable machine learning join efforts toward personalised medicine in a novel decision-making tool (2022). arXiv:2205.06234.

[24] J. H. Friedman, B. E. Popescu, Predictive learning via rule ensembles, The annals of applied statistics (2008) 916–954.
22